\documentclass{llncs}

\usepackage{times}
\usepackage{latexsym}
\usepackage{multirow}
\usepackage{booktabs}
\usepackage{nicefrac}
\usepackage{flushend}
\usepackage{adjustbox}
\usepackage{enumitem}
\usepackage{epigraph}
\usepackage{hyperref}
\usepackage{minibox}
\usepackage[misc]{ifsym} 
\usepackage[normalem]{ulem} 
\usepackage[usenames, dvipsnames]{color}

\newcommand{\under}[1]{\underline{\smash{#1}}}

\newcommand{\squishlist}{
 \begin{list}{$\bullet$}
  { \setlength{\itemsep}{0pt}
     \setlength{\parsep}{1pt}
     \setlength{\topsep}{1pt}
     \setlength{\partopsep}{0pt}
     \setlength{\leftmargin}{1.5em}
     \setlength{\labelwidth}{1em}
     \setlength{\labelsep}{0.5em} } }
 \newcommand{\squishend}{\end{list}}
 
\newtheorem{task}{Task}
\author{Paramita Mirza$^1$ (\Letter), Simon Razniewski$^1$, Fariz Darari$^2$, Gerhard Weikum$^1$}

\institute{$^1$Max Planck Institute for Informatics, Saarbr\"ucken, Germany\\
\texttt{\{paramita, srazniew, weikum\}@mpi-inf.mpg.de}
\\$^2$Universitas Indonesia, Depok, Indonesia\\
\texttt{fariz@cs.ui.ac.id}}

\title{
Enriching Knowledge Bases with Counting Quantifiers
}

\begin{document}

\maketitle

\begin{abstract}
Information extraction traditionally focuses on extracting relations between identifiable entities, such as {\em $\langle$Monterey, locatedIn, California$\rangle$}. Yet, texts often also contain Counting information, stating that a subject is in a specific relation with a number of objects, without mentioning the objects themselves,
for example, \textit{``California is divided into 58 counties''}. Such counting quantifiers can help in a variety of tasks such as query answering or knowledge base curation, but are neglected by prior work.

This paper develops the first full-fledged system for extracting counting information from text, called \textit{CINEX}. We employ distant supervision using fact counts from a knowledge base as training seeds, and develop novel techniques for dealing with several challenges: 
\emph{(i)} non-maximal training seeds due to the incompleteness of knowledge bases, 
\emph{(ii)} sparse and skewed observations in text sources, and
\emph{(iii)} high diversity of linguistic patterns.
Experiments with five human-evaluated relations show that CINEX can achieve 60\% average precision for extracting counting information. In a large-scale experiment, we demonstrate the potential for knowledge base enrichment by applying CINEX to 2,474 frequent relations in Wikidata. CINEX can assert the existence of 2.5M facts for 110 distinct relations, which is 28\% more than the existing Wikidata facts for these relations.
\end{abstract}

\section{Introduction}




\subsubsection {Motivation.} 
General-purpose knowledge bases (KBs)
like Wikidata, DBpedia or YAGO \cite{vrandevcic2014wikidata,dbpedia,suchanek2007yago} find increasing use in applications such as question answering, entity search or document enrichment, and their automated construction from Internet sources has been greatly advanced. 
So far, information extraction (IE) to this end has focused on fully qualified subject-predicate-object (SPO) facts such as
{\em $\langle$Monterey, locatedIn, California$\rangle$}.
However, texts often contain only 
counting
information: the number of objects that stand in a specific relation with a certain entity, without mentioning the objects themselves.
Examples are: \emph{``California is divided into 58 counties''},
\emph{``Clint Eastwood directed more than twenty movies''} or
\emph{``Trump has three sons and two daughters''}.

This kind of knowledge can be codified into an extension of
existentially quantified formulas known in AI and logics as
{\em counting quantifiers (CQs)}: they assert the existence of
a specific number of SPO triples without fully knowing the
triples themselves.
Counting information can substantially
extend the scope and value of knowledge bases. 
First, they allow accurate answers for queries that involve counts 
(e.g., number of counties per US state)
or existential quantifiers (e.g., directors who made at least 5 movies). 
%
Second, an important use case is KB curation
\cite{Dong:PVLDB2014,Tan:WSDM2014}.
KBs are notoriously incomplete, contain erroneous triples,
and are limited in keeping up with the pace of real-world changes.
Counting 
information helps to identify gaps and inaccuracies.
For example, knowing the exact number of counties in California 
or a lower bound for the number of films directed by Eastwood
are important cues to complete and enrich a KB.
%

\vspace{-7pt}
\subsubsection{State-of-the-Art and Challenges.} 
The predominant approach to extracting facts for KB population
is distant supervision, using seeds for the SPO triples
of interest (e.g., \cite{Mintz:ACL2009,Suchanek:WWW2009}). 
The seeds are usually taken from an initial KB
or are manually compiled. Spotting the seeds
in a text corpus (e.g., {\em Clint Eastwood}, {\em directed}
and {\em Gran Torino})
then allows learning patterns for relations (e.g., \textit{``director of''}
or \textit{``$\langle${\em someone}$\rangle$'s masterpiece''}),
which in turn lead to observing new fact candidates.
This methodology is known as the pattern-relation duality principle
\cite{Brin:WebDB1998}.

Distant
supervision is a natural approach
for extracting 
counting information as well: the cardinality of distinct O 
arguments for a given SP pair, $n:=|\{O\,|\,SPO \in \mathit{KB}\}|$, serves as a seed for
the 
counting assertion, $\langle S, P, \exists n\rangle$.
However, it is more challenging than traditional
SPO-fact extraction and needs to cope with several
issues:
\squishlist
\item[1)] \emph{Non-maximal seeds:} Unlike for
SPO-fact extraction, the incompleteness of KBs not only leads to a reduction in the number of seeds, but to seeds that systematically 
underestimate the
count of facts 
that are valid in reality. 
For example, a KB that knows only a subset of Trump's children, say
three out of five, leads
to a non-maximal seed that may reward spurious patterns like \textit{``owns \under{three} golf resorts''} at the cost of patterns like \textit{``his \under{five} children''}. 
Even worse, KBs often have complete blanks on certain relations, e.g.,
not knowing any of Eastwood's\linebreak movies despite labeling his occupation as
{\em film director} and {\em film producer} \mbox{\ \ \ \ \ \ } \linebreak
({\footnotesize{\url{https://www.wikidata.org/wiki/Q43203}}}).

\item[2)] \emph{Sparse and skewed observations:} For many relations, 
counting
information
is expressed in text in a sparse and highly skewed way. 
For example, the non-existence of children is rarely mentioned.
For musicians, the first Grammy someone has won often has more
mentions than later ones, hence giving undue weight to the pattern \textit{``his/her first award''}. The number of members in a music band is often around four, 
which makes it hard to learn patterns for very large or very small bands.

\item[3)] \emph{Linguistic diversity:} 
Counting
information can be expressed in a variety of linguistic forms like\\
\textit{(i)} {\em explicit numerals} as cardinal numbers 
(e.g., \textit{{``has \under{five} children''}}),\\
\textit{(ii)} \emph{lower bounds} via ordinal numbers (e.g., \textit{{``her \under{third} husband''}}),\\
\textit{(iii)} {\em number-related noun phrases} such as
\textit{`twins'} or \textit{`quartet'},\\ 
\textit{(iv)} {\em existence-proving articles} as in
\textit{``has \under{a} child''},\\
\textit{(v)} {\em non-existence adverbs} such as \textit{`never'} and \textit{`without'}.
\squishend
%


Open IE methods 
\cite{Mausam:IJCAI2016}
cannot cope with these challenges.
For example, the sentence \textit{``Trump has five children''} would 
typically result in the triple {\em $\langle$Trump, has, five children$\rangle$}, failing to recognize that `five' is a numeric modifier of `children'.
On the other hand, IE methods with pre-specified relations
for KB population (e.g., NELL \cite{NELL:Mitchell2015}) 
capture relevant O values only for few relations specified to have
numeric literals as their range, such as {\em numberofkilledinbombing} 
or {\em earthquakecasualitiesnumber}\linebreak
({\footnotesize{\url{http://rtw.ml.cmu.edu/rtw/kbbrowser/}}}).



\subsubsection{Approach and Contributions.} 
In this paper, we develop the first 
full-fledged system for
\underline{C}ounting \underline{In}formation \underline{Ex}traction,
called CINEX.
Our method is based on 
machine learning for
sequence 
labeling,
judiciously designed to cope with the outlined challenges.
We leverage distant supervision from fact counts in a given KB,
but devise special techniques to handle non-maximal seeds, 
sparseness and skew in observing count information in text, 
and linguistic diversity of patterns.
We counter non-maximal seeds
(Challenge 1) by relaxing matching conditions for numbers higher than KB counts, and by reducing the training to popular, more complete entities.
Sparseness and skew (Challenge 2) are addressed by 
discounting uninformative numbers using entropy measures.
Linguistic variance (Challenge 3) is handled by careful consolidation of detected mentions. 
We 
devise both a traditional feature-based conditional random field (CRF) 
and a bi-directional LSTM-CRF model using TensorFlow, finding that both perform roughly comparable, although the traditional approach is more robust when dealing with noisy training data.
%

The salient original contributions of this paper are:
\squishlist
\item The 
methodology of our extraction system, 
CINEX.
\item An empirical evaluation with five manually annotated relations,
showing 60\% precision on average.
\item An application and large-scale experimental study of 
CINEX
on 2,474 frequent relations of Wikidata,
showing that 
counting
information can extend the SPO facts in 
Wikidata for 110 distinct relations by 28\%.
\item Code and data made available to the research community on Github.\footnote{\url{https://github.com/paramitamirza/CINEX}}
\squishend

The remainder of this paper is structured as follows. 
In Section~\ref{sec:formalization} we specify the scope of 
counting quantifiers
and discuss the incompleteness of
KBs, using Wikidata as a reference point.
Section~\ref{sec:approach} presents our methodology for
extracting 
counting information at large scale, which we then detail in Sections~\ref{sec:identify-mentions} and~\ref{sec:consolidation}.
Section~\ref{sec:eval} gives experimental results on the quality of
our extraction method, with a particular focus on how 
CINEX can enrich the Wikidata KB in Section~\ref{sec:broad}.
Section~\ref{sec:relatedwork} discusses related work.

\section{Counting Information in Knowledge Bases}
\label{sec:formalization}




Counting quantifiers
for a KB with SPO triples are statements on a subset of the SPO arguments.
We focus on the dominant case of quantification of O arguments for a given SP pair.
We write 
counting
statements 
as $\langle S, P, \exists n \rangle$, where $S$ is the subject, $P$ is the predicate and $n$ is a natural number (including zero). For instance, the statement that President Garfield has 7 children would be written as $\langle \mathit{Garfield, hasChild}, \exists 7 \rangle$. In the OWL description logics, 
this statement is written as:

\vspace{4pt}
\noindent\texttt{\footnotesize{ClassAssertion(ObjectExactCardinality(7 :hasChild) :Garfield)}}

\vspace{-7pt}
\subsubsection{Wikidata.}
\begin{figure}[t]
\begin{center}
\includegraphics[width=0.55\textwidth]{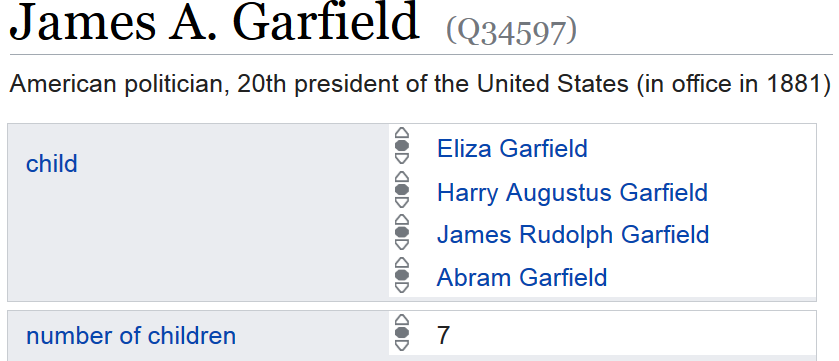}
\end{center}
\caption{SPO facts and counting information in Wikidata.}
\label{fig:garfield}
\end{figure}

To illustrate how today's KBs deal with 
counting information, we briefly discuss the 
case of Wikidata, presumably the world's largest and best curated
publicly available KB. Wikidata already contains 
counting
relations for a few topics such as {\em numberOfChildren}, {\em numberOfSeasons} (of a TV series), or {\em numberOfHouseholds} (of an administrative entity). This 
information can coexist with fully qualified SPO facts. 
Regarding children, for example, Wikidata knows 4 out of the 7 children of President Garfield by name, and knows that he had 7 in total (see Fig.~\ref{fig:garfield}). However, the {\em numberOfChildren} predicate is asserted for only 0.2\% of persons in Wikidata so far. Even the \emph{child} property is asserted for only 2.2\% of persons, creating uncertainty about whether the others have no children or whether Wikidata does not know about them.

%
%


Counting information is beneficial for search and
question answering, for example to answer \textit{``Which US presidents were married twice?''}
We analyzed the number of questions in the 
TREC 2003, 2004 and 2007 QA test datasets~\cite{trec2007}, and
found that 5\% to 10\% of the questions (typically starting with \emph{``How many''}) fall into this category.



\vspace{-10pt}
\subsubsection{Potential for KB Enrichment.} 
To quantitatively assess the gap in Wikidata, for which 
counting
information can contribute
to KB enrichment, we had one expert read the Wikipedia articles of 200 randomly selected people,
with the task of comparing the text-borne 
counting information on 
the \emph{hasChild} relation with the explicitly stated children names.
The expert was instructed to look at two kinds of cues: 
\emph{i)} \textit{explicit numerals} expressing 
counting information,
\emph{ii)} \textit{counting names} of children mentioned in the article.
We compare these numbers against 
\emph{iii)} the \textit{Wikidata SPO triples} for the person's {\em hasChild} predicate.
Note that approach ii) corresponds to what standard IE aims to achieve
(i.e., extracting full triples and then counting).

%
We found that
counting 
information via numerals allows the discovery of children counts for 12\% of all test entities, while names of children are only mentioned for 7\%, and Wikidata contains 
facts about children for only 2.5\%.
As for the total number of children,
counting
information 
asserts the existence
of \emph{twice} as many children, 
i.e., 0.35 children per person, 
as spotting and counting children names (0.18), and even \emph{eleven} times more than
Wikidata currently knows of (0.03).

\section{System Overview}
\label{sec:approach}

\newtheorem{myproblem}{Problem}
The CINEX system aims to solve the following problem:
\begin{myproblem}[Counting Quantifier Extraction]
Given a text about a subject $S$, and a predicate $P$, the task of counting quantifier (CQ) extraction is to determine the number of objects with which $S$ stands in relation regarding $P$.
\end{myproblem}
For instance, given the sentence \emph{``Trump has three sons and two daughters''}, the output for the predicate \textit{numberOfChildren} should be 5.

\begin{figure*}[bt]
\begin{center}
\includegraphics[width=0.8\textwidth]{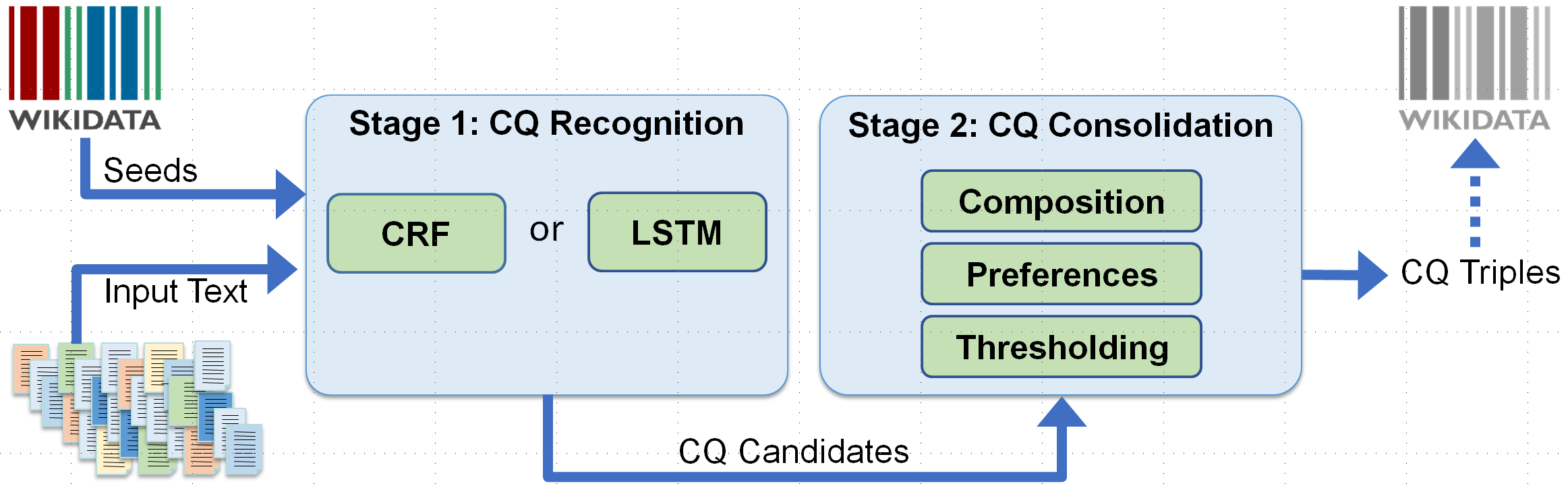}
\end{center}
\caption{Overview of the CINEX system.}
\label{fig:cinex-overview}
\end{figure*}

Figure \ref{fig:cinex-overview} gives a pictorial overview of the system architecture of CINEX. We split the overall task into two main components: the recognition of counting information and the consolidation of intermediate results into the final output of counting quantifiers. 
These components are presented in Sections
\ref{sec:identify-mentions} and \ref{sec:consolidation}, respectively.

CINEX utilizes seeds from Wikidata in a judicious way in order to train a model for CQ recognition, using one of two options: a conditional random field (CRF) or a
bidirectional LSTM neural network.
When applied to new text, the output of the
recognition model is a set of CQ candidates,
which are often fairly noisy, though.
Subsequently, the second stage of CINEX -- CQ consolidation -- cleans and aggregates the
counting information and produces the final
output of CINEX. 
The resulting CQ triples could potentially be
added to a knowledge base such as Wikidata.

\section{Counting Quantifier Recognition}
\label{sec:identify-mentions}

The first stage of CINEX aims to recognize counting information in text,
this way collecting a pool of CQ candidates for further cleaning and consolidation.
We cast the CQ recognition into a sequence labeling task, operating on 
a per-sentence basis and learned separately for each predicate $P$. 
We are interested in counting information for a given
subject-predicate (SP) pair and assume that the subject is already identified by the
sentence context (e.g., the main entity featured in a document, like a Wikipedia article
about S or S's homepage on the Web).
Furthermore, we assume that the input sentence is pre-processed by detecting terms
that indicate counting information: cardinals, ordinals and number-related terms (numterms).

\begin{task}[Counting Quantifier Recognition]
Given a sentence about subject $S$ and predicate $P$
containing at least one cardinal, ordinal
or number-related term (numterm),
the task of CQ recognition is to label each token of the sentence
with one of the following tag: (i) \textsc{count},  for denoting a CQ mention, 
(ii) \textsc{comp}, for denoting compositional cues and (iii) \textsc{o}, for others.
\end{task}

The following shows an example:
\vspace{-10pt}
\begin{center}
\begin{adjustbox}{width=\textwidth}
\begin{tabular}{l| c c c c c c c c c c c c c c c }
sentence & \textit{Jolie} & \textit{brought} & \textit{her} & \textit{\under{twins}} & , & \textit{\under{one}} & \textit{daughter} & \textit{and} & \textit{\under{three}} & \textit{adopted} & \textit{children} & \textit{to} & \textit{the} & \textit{gala} & \textit{.} \\
pre-processed & \textit{Jolie} & \textit{brought} & \textit{her} & \textsc{numterm} & , & \textsc{cardinal} & \textit{daughter} & \textit{and} & \textsc{cardinal} & \textit{adopted} & \textit{children} & \textit{to} & \textit{the} & \textit{gala} & \textit{.} \\
output tags & O & O & O & \textsc{count} & \textsc{comp} & \textsc{count} & O & \textsc{comp} & \textsc{count} & O & O & O & O & O & O \\ 
\end{tabular}
\end{adjustbox}
\end{center}

\subsubsection{Sequence Labeling Models.}
Our problem resembles the Named Entity Recognition (NER) task, with Conditional Random Fields (CRFs) being a typical choice of sequence labeling models. 
In order to generalize patterns beyond specific numeric values/tokens, we pre-process sentences
to lift these specific tokens into placeholders {\em cardinal}, {\em ordinal} and {\em numeric term (numterm)}. 
For instance, the sentence 
\textit{``Donald Trump has} \textit{three} \textit{children from his} \textit{first} \textit{wife.''} 
becomes \textit{``Donald Trump has} \textsc{cardinal} \textit{children from his} \textsc{ordinal} \textit{wife.''} 

CINEX learns one sequence labeling model for each predicate of interest
(e.g., with separate models for children and spouses).
We have devised solutions based on two sequence labeling methods:
\vspace{-5pt}
\begin{enumerate}
\item \emph{Feature-based model.} We constructed a CRF-based sequence classifier using CRF++ \cite{kudo2005crf++} with n-gram features (up to pentagrams), taking into account lemmas and placeholders (e.g., \{\textit{Trump}, \textit{have}, \textsc{cardinal}, \textit{child}, \textit{from}\}) instead of the original tokens. 
\item \emph{Neural model.} We adopt the bidirectional LSTM-CRF architecture proposed in~\cite{N16-1030} using TensorFlow, presently the state-of-the-art method for sequence-to-sequence learning, 
to build our sequence labeling model. The neural architecture takes into account words, placeholders and character embeddings to represent the input sequence. The neural model should be able to exploit, for example, that word embeddings for \textit{`children'}, \textit{`daughters'} and \textit{`sons'} are close to each other in the embedding space. Furthermore, word embeddings for 
out-of-vocabulary words such as \textit{`ennealogy'} can be generated via character embeddings, recovering similarity to e.g.\ \textit{`pentalogy'}.
\end{enumerate}




\vspace{-15pt}
\subsubsection{Incompleteness-Aware Distant Supervision.}
We employ distant supervision to generate training data, as common in relation extraction~\cite{distant-first,Mintz:ACL2009,Suchanek:WWW2009}.
Given a knowledge base (KB) relation $P$, 
for each entity $S$ in the KB that appears as the subject of $P$,
we retrieve \emph{(i)} the \textit{triple count} $|\langle S, P, * \rangle|$ from the KB and \emph{(ii)} sentences about $S$ containing \textit{candidate mentions}, e.g., cardinal numerals. Candidate mentions that are equal to or representing the triple count will be 
labelled with the tag \textsc{count} denoting counting quantifier mentions,  
i.e., as positive examples. 
Otherwise, candidate mentions will be labeled with the 
$O$ tag,
i.e., as negative examples, like any other non-candidate mentions (e.g., non-numerals). We built separate training data for each relation $P$ of interest.

Incomplete information from the KB used as the ground truth may negatively affect the quality of training data resulting from the distant supervision approach. To mitigate the effect that KB incompleteness has on training data quality, 
we investigated filtering
the ground truth based on \textit{subject popularity}, according to the number of stored KB triples for that subject, 
which is also highly correlated with other popularity measures like PageRank or Wikipedia article length.
For example, for 10 random entities from the 99th, 90th and 80th percentile wrt.\ popularity, the mean difference between Wikidata children counts and a manually established ground truth from Wikipedia is 0.8, 1.5 and 2.4, respectively.
Assuming that popularity and completeness are correlated in general, we can thus trade training data quantity for quality by disregarding less popular entities during training.





Candidate counts that are higher than the KB count are normally considered as not expressing the object count for the relation of interest, i.e., as negative training examples. \emph{But this can also happen to mentions that actually express the correct count}, when the KB is incomplete and only knows counts lower than the correct one.
Our remedy is to treat mentions higher than KB counts neither as positive nor as negative examples, but to simply exclude them from the training set. 
However, there is the need to maintain enough negative examples; otherwise, the classifier would get overly optimistic.
For this purpose we utilize upper bound information of triple counts specific to each relation, i.e., the triple count at 99th percentile 
(e.g., 3 for number of spouses), 
as found in the KB. A higher count mention will then still be treated as a negative example if it is deemed to be impossible to represent count information for the relation in question.

Furthermore, the more frequent a certain number occurs in a text, the more probable it is to occur in various contexts. As a way to give the classifier less noisy training examples, we ignore sentences that contain count mentions of numbers that have a low entropy in the given text, even when they represent the actual object count. This way we ensure that the models only learn from correct number mentions in the right context.


\vspace{-4pt}
\subsubsection{Linguistic Diversity.}
As mentioned in the introduction, there are several ways to express count information in natural language text, cardinals and ordinals being only the most obvious ones. 

\vspace{-4pt}
\paragraph{Number-related terms.}
We exploited the \textit{relatedTo} relation in ConceptNet \cite{SPEER12.1072} for collecting around 1,200 terms related to numbers. 
The terms are split into two groups, those having Latin/Greek prefixes\footnote{\url{http://phrontistery.info/numbers.html}} and those not having them. For the first group, we generated a list of Latin/Greek prefixes (e.g., \textit{quadr-}) and a list of possible suffixes (e.g., \textit{-plets}). When generating training data, a term with Latin/Greek affixes was labeled with 
the positive \textsc{count} tag
if its prefix matched the triple count. 
For feature-based models we also replaced such terms in the input with placeholders \textsc{numterm} appended with their Latin/Greek suffixes, while we use the original tokens for neural models.

From the second group we manually selected 15 terms that were especially strongly associated with specific counts (e.g., \textit{twins}, \textit{dozen}). 
During preprocessing, these terms 
are then either replaced with corresponding terms/phrases containing cardinal numbers, e.g., 
\textit{thrice} $\rightarrow$ \textit{three times}
 and \textit{a dozen} $\rightarrow$ \textit{twelve}, or replaced with corresponding Latin/Greek suffix placeholders (e.g. 
 \textsc{numterm-plets} for \textit{twins}).

\vspace{-6pt}
\paragraph{Indefinite articles.}

Indefinite articles (i.e., \textit{`a'}, \textit{`an'})
are similar to the ordinal \emph{first} insofar as they can express the existence of at least one object. We initially planned to treat them this way, yet due to their overwhelming frequency our classifiers could not cope with them. Thus we now disregard them in the training stage and only consider them as candidate mentions when applying the learned models, by replacing them with the \textsc{cardinal} placeholder, and treating them as the mention \emph{one}.




\vspace{-6pt}
\subsubsection{Compositionality.}


To account for compositional mentions occurring in one sentence,
we introduce 
an extra label, \textit{compositionality tag} (\textsc{comp}), 
for the sequence labeling models.
During training data generation, we identify consecutive candidate tokens 
with label \textsc{count} such that
 \textit{(i)} the sum of their values is equal to the triple count 
 and \textit{(ii)} there exist \textit{compositional cues} (commas and \textit{`and'}) in between, which are then tagged with 
the \textsc{comp} label.

\section{Counting Quantifier Consolidation}
\label{sec:consolidation}




Once tokens expressing counting or compositionality information have been identified, these need to be consolidated into a single prediction for the number of objects.

\begin{task}[Counting Quantifier Consolidation]
For a given subject $S$ and predicate $P$, the input to this second stage is
a set of token lists, where each token list consists of words/numbers and their
corresponding input and output labels (i.e., cardinal, ordinal, numterm, count or comp)
and at least one token is tagged cardinal, ordinal or numterm.
The desired output is a single number for the counting quantifier for $S$ and $P$,
that is, the correct number of objects for $S$ and $P$.
\end{task}

For example, 
for the pair $\langle$\textit{AngelinaJolie}, \textit{hasChild}$\rangle$, the 
following token lists may have
been detected (annotated as \under{counting information} and $\lbrack$compositional cues$\rbrack$, with confidences as subscripts):
\begin{enumerate}[label=$l_\arabic*$]
\item \textit{Angelina has a grand total of \under{six}$_{0.4}$ children together : \under{three}$_{0.3}$ biological $\lbrack$and$\rbrack_{0.6}$ \under{three}$_{0.5}$ adopted .}
\item  \textit{The arrival of the \under{first}$_{0.5}$ biological child of Jolie and Pitt caused an excited flurry with fans .}
\item \textit{On July 12 , 2008 , she gave birth to \under{twins}$_{0.8}$ : \under{a}$_{0.1}$ son , Knox Léon , $\lbrack$and$\rbrack_{0.5}$ \under{a}$_{0.2}$ daughter , Vivienne Marcheline .}
\end{enumerate}

We use the following algorithm to consolidate the counting quantifier (CQ) candidates from
these labeled token lists.

\newtheorem{algo}{Algorithm}

\begin{algo}[Mention Consolidation] \mbox{}
\vspace{-0.15cm}
\begin{enumerate}
\item \emph{Sum up compositional mentions.} Mentions having compositional cues in between are summed up,
and their confidence score is set to the highest confidence score of the mentions.
\item \emph{Select prediction per type.} For multiple mentions of type cardinal and number-related term, only the mention with the highest confidence is retained if it is above a certain threshold, with compositional mentions treated like cardinals. For ordinals, we always select the highest ordinal available in the candidate pool,
regardless of the confidence scores.
\item \emph{Rank mention types.} In the last step, the final prediction is chosen based on the preference $n_{cardinal} \gg n_{numterm} \gg n_{ordinal} \gg n_{article}$, i.e., whenever a cardinal mention exists, it is returned as final answer, otherwise a number-related term, ordinal or article.
\end{enumerate}
\end{algo}
In the example above, in the first step, the two mentions of \under{\em three} in $s_1$ are summed up to one mention \under{\em 6}$_{0.5}$, and the two
indefinite articles in $s_3$ are combined into \under{\em 2}$_{0.2}$. In the second step, \under{\em 6}$_{0.5}$ is chosen as highest-confidence cardinal, \under{\em twins}$_{0.8}$ as highest ranking numterm (with numerical value 2), and \under{\em first}$_{0.5}$ as highest ranking ordinal. In the last step, the cardinal \under{\em 6}$_{0.5}$ or the term \under{\em twins}$_{0.8}$ is chosen as final prediction, depending on whether the confidence threshold is below 0.5 or not.


\vspace{-10pt}
\subsubsection{Confidence Scores.} We interpret \textit{marginal probabilities} given by CRFs, i.e., the probability of a token labeled with a certain tag resulting from forward-backward inference, as the \textit{confidence scores} of identified mentions. When a CRF layer is not applied on top of the neural models, the probabilities are simply given by the \textit{softmax} output layer.

\vspace{-10pt}
\subsubsection{Count Zero.}

We so far only considered counting information for counts greater than zero. Reliably recognizing subjects without objects is difficult for two reasons, (i) because reliable training data is even harder to come by, and (ii) because the count zero is neither expressed via cardinals nor ordinals or indefinite articles. We thus consider count zero only in passing, focusing on two especially frequent ways to express it: \textit{(i)} determiners \textit{`no'} and \textit{`any'} (used in negation) and \textit{(ii)} non-existence-proving adverbs \textit{`without'} and \textit{`never'}.
We approach their labeling in a manner similar to the identification of count information via indefinite articles,
i.e., not using the count quantifier cues for training but considering them when applying the models. 

We performed text preprocessing beforehand to ensure that the non-existence cues can be discovered by the learned models.
This preprocessing step includes transforming sentences 
containing \textit{`not-any'}, \textit{`never'} and \textit{`without'} into sentences containing \textit{`no'} and \textit{`0'}, for example:
\vspace{-5pt}
\begin{center}
\small
\begin{tabular}{ l l l }
 \textit{They did\under{n't} have \under{any} children} & $\rightarrow$ &  \textit{They have \under{no} children} \\  
 \textit{He has \under{never} been married} & $\rightarrow$ & \textit{He has been married \under{0} times} \\
 \textit{The marriage was \under{without} children} & $\rightarrow$ & \textit{The marriage was with \under{no} children}. \\
\end{tabular}
\end{center}
Finally, textual occurrences of \textit{`no'} and \textit{`0'} are replaced with \textsc{cardinal} and treated as count zero.



\section{Experiments}
\label{sec:eval}


\subsection{Experimental Setup}
\label{subsec:eval-baseline}

\subsubsection{Dataset.} 
We chose Wikidata as our source KB and Wikipedia pages about given subject entities as our source text for the distant supervision approach.\footnote{Both in their version as of March 20, 2017.}
While some Wikidata properties are self-explanatory, like \textit{child} or \textit{spouse}, some others are overloaded, i.e., used in highly diverse domains with different semantics depending on the type of the subject entities, e.g. \textit{has part}.
Thus, we define \textit{relations} in our experiments as pairs of a Wikidata subject type/class and a Wikidata property.
We focus on five diverse relations (listed in Table~\ref{tbl:predicates} under the \textit{Relation} column) using the four Wikidata properties already used in~\cite{MirzaRDW17}, but using two specific Wikidata classes for the overloaded \textit{has part} property, i.e., \textit{series of creative works} 
and \textit{musical ensemble}. 
%
%
\begin{table*}[t!]
\centering
\begin{adjustbox}{width=\textwidth}
\begin{tabular}{ll|l|r}
\toprule
 Wikidata subject class  & Wikidata property & Relation & \#Subjects \\ 
\midrule
series of creative works (Q7725310) & has part (P527) & containsWork & 642 \\
musical ensemble (Q2088357) & has part (P527) & hasMember & 8,901 \\
admin.\ territ.\ entity (Q56061) & contains admin.\ territ.\ entity (P150) & containsAdmin & 6,266 \\
human (Q5) & child (P40) & hasChild & 40,145 \\
human (Q5) & spouse (P26) & hasSpouse & 45,261 \\
\bottomrule
\end{tabular}
\end{adjustbox}
\caption{Number of Wikidata instances as subjects (\#Subject) of each relation in the training set.
}
\label{tbl:predicates}
\end{table*}
%
We use 
four
sets of entities for training and evaluation:
\begin{enumerate}
\item \emph{Training set}: For each relation, all subject entities with an English Wikipedia page that have at least one object in Wikidata, except those used for development and testing (counts are shown in Table~\ref{tbl:predicates}).
\item \emph{Manual test set}: 200 entities per relation randomly chosen from the training set (i.e., have at least one object).
\item \emph{Automated test set}: 200 of the 10\% most popular entities per relation removed from the training set (i.e., have at least one object).
\item \emph{Zero-count test set}: 64 and 168 entities for the \textit{hasChild} and \textit{hasSpouse} relations, respectively, which are entities in Wikidata having child (P40) and spouse (P26) properties set to the special value \emph{no-value}.
\end{enumerate}
%
%
For the \textit{manual test set} we manually annotated mentions in text that correspond to counting quantifiers, and established the correct object count from Wikipedia.
%
The \textit{automated test set} is used for parameter tuning of the neural models, and as silver standard for evaluating our system beyond the 5 gold-annotated relations.
%
%
For evaluating zero-count quantifier detection, we use two relations for which manually created data from Wikidata is available.

\vspace{-10pt}
\subsubsection{Hyperparameters.} We set 0.1 as the confidence score threshold in the mention consolidation task (Section~\ref{sec:consolidation}), after experimenting with varying values. For training the neural models, we employed Adam~\cite{Adam} with a learning rate of 0.001. Using stochastic gradient descent (SGD) with a gradient clipping of 5.0 as reported in~\cite{N16-1030} results in worse performance. The LSTM network uses a single layer with 300 dimensions. The hidden dimension of the forward and backward character LSTMs are 100. We set the dropout rate to 0.5. We also use GloVe pre-trained embeddings~\cite{pennington2014glove} to initialize our lookup table.

\subsection{Evaluation}
\label{sec:auto-manual-eval}

\subsubsection{Evaluation Scheme.} 
We evaluate our system, CINEX (\under{C}ounting \under{In}formation \under{Ex}tract\-ion), on quantifier recognition, quantifier consolidation, and on the end-to-end task with the following metrics:

We use \textit{precision}, \textit{recall} and \textit{F1-score} to evaluate how well the system can identify counting information in a given text.
For entities for which the system recognized at least one counting quantifier (CQ) candidate, we then measure \textit{precision} in choosing the correct final CQ.
Finally, we evaluate the system for the end-to-end task in terms of \textit{coverage}, i.e., for how many subject entities the system can extract correct object counts from text, and 
\textit{Mean Absolute Error (MAE)},
to understand how much system predictions deviate from the truth.

\vspace{-10pt}
\subsubsection{Quantifier Recognition. }

\begin{table}[t]
\begin{center}
\setlength\tabcolsep{3pt}
\begin{tabular}{@{}l|rrr|rrr|rrr|rrr}
\toprule
\multirow{3}{*}{Relation} & \multicolumn{3}{c|}{\multirow{2}{*}{Baseline~\cite{MirzaRDW17}}} & \multicolumn{9}{c}{CINEX} \\
 &  &  &  &  \multicolumn{3}{c|}{CRF} & \multicolumn{3}{c|}{biLSTM} & \multicolumn{3}{c}{biLSTM-CRF} \\
 & P & R & F1 & P & R & F1 & P & R & F1 & P & R & F1 \\ 
\midrule
containsWork  &  22.4  &  24.0  &  23.1  &  61.9  &  29.3  &  \textbf{39.8}  &  61.1  &  19.6  &  29.6  &  54.9  &  28.9  &  37.8 \\
hasMember  &  1.5  &  4.3  &  2.2  &  55.7  &  56.5  &  \textbf{56.1}  &  38.2  &  18.8  &  25.2  &  35.9  &  33.3  &  34.6 \\
containsAdmin  &  51.1  &  64.3  &  57.0  &  72.5  &  82.9  &  77.3  &  78.4  &  82.9  &  80.6  &  78.7  &  84.3  &  \textbf{81.4} \\
hasChild  &  6.4  &  49.4  &  11.4  &  54.5  &  44.4  &  \textbf{49.0}  &  33.9  &  11.7  &  17.4  &  26.1  &  14.8  &  18.9 \\
hasSpouse  &  1.9  &  12.1  &  3.3  &  58.2  &  67.2  &  \textbf{62.4}  &  20.4  &  36.2  &  26.1  &  27.1  &  32.8  &  29.7 \\
\bottomrule
\end{tabular}
\end{center}
\caption{Performance of CINEX on recognizing counting quantifier mentions, with different architectures and in comparison with the baseline. Highest F1-score per relation in boldface.}
\label{tbl:identify-mentions-eval}
\end{table}

\begin{table}[t]
\begin{center}
\setlength\tabcolsep{3pt}
\begin{tabular}{@{}lrrr|rrr|rrr|rrr}
\toprule
\multirow{3}{*}{Relation} & \multicolumn{3}{c|}{Baseline~\cite{MirzaRDW17}} & \multicolumn{9}{c}{CINEX-CRF (per type)} \\
  & \multicolumn{3}{c|}{Cardinals} & \multicolumn{3}{c|}{Cardinals} & \multicolumn{3}{c|}{Numt.+Art.} & \multicolumn{3}{c}{Ordinals} \\
 & P & R & F1 & P & R & F1 & P & R & F1 & P & R & F1 \\ 
\midrule
containsWork & 22.4 & 77.8 & \textbf{34.8} & 60.0 & 18.3 & 28.1 & 53.1 & 98.1 & 68.9 & 77.6 & 19.9 & 31.7 \\
hasMember & 1.5 & 25.0 & 2.9 & 50.0 & 33.3 & \textbf{40.0} & 55.7 & 64.2 & 59.6 & 100 & 25.0 & 40.0 \\
containsAdmin & 51.1 & 64.3 & 57.0 & 84.1 & 82.9 & \textbf{83.5} & 0 & 0 & 0 & 0 & 0 & 0 \\
hasChild & 6.4 & 72.7 & 11.8 & 75.6 & 56.9 & \textbf{64.9} & 24.3 & 100 & 39.1 & 7.7 & 2.3 & 3.5 \\
hasSpouse & 1.9 & 87.5 & 3.7 & 76.9 & 90.9 & \textbf{83.3} & 0 & 0 & 0 & 85.3 & 63.0 & 72.5 \\
\bottomrule
\end{tabular}
\end{center}
\caption{Performance of CINEX-CRF on recognizing counting quantifier mentions, per mention type. \textit{Numt.} stands for number-related terms, \textit{Art.} for indefinite articles. Baseline comparison is only for cardinals (highest F1-score per relation in boldface).}
\label{tbl:identify-mention-types-eval}
\end{table}

We report in Table~\ref{tbl:identify-mentions-eval} the performance results of different architectures wrt.\ precision, recall and F1-score. We also compare our system with the best 
performing method for extracting cardinals reported in~\cite{MirzaRDW17} as baseline.
As one can see, feature-based CRF models are the most robust sequence labeling approach across relations for this task, although the neural models achieve 
higher
F1-score with 3.3 percentage point difference for \textit{containsAdmin}. Adding a CRF layer on top of bidirectional LSTM models improves performance across relations, although this architecture still fails to beat the 
feature-based CRF models
in most cases. We conjecture that this is due to neural models being much more prone to overfitting to noisy distantly supervised training data. Still, both feature-based and neural models consistently outperform the baseline by a large margin, in particular wrt.\ precision.



In Table~\ref{tbl:identify-mention-types-eval} we split this analysis further by mention type. This provides a more fair comparison with the baseline that only considers cardinal numbers. Still, CINEX-CRF achieves a higher precision on all relations, and a higher F1-score on 4 out of 5. We also see variety within the mention types and relations, ordinals for instance being well picked up for \emph{hasSpouse}, but badly for \emph{hasChild}.


\vspace{-10pt}
\subsubsection{Quantifier Consolidation. }

\begin{table}[t!]
\begin{center}
\setlength\tabcolsep{3pt}
\begin{tabular}{@{}lrrr|rrr|rr|rr|rr}
\toprule
\multirow{2}{*}{Relation} & \multicolumn{3}{c|}{\multirow{2}{*}{Baseline~\cite{MirzaRDW17}}} & \multicolumn{3}{c|}{\multirow{2}{*}{CINEX-CRF}} & \multicolumn{6}{c}{CINEX-CRF (per type)} \\
 & & & & & & & \multicolumn{2}{c|}{Cardinals} & \multicolumn{2}{c|}{Numt.+Art.} & \multicolumn{2}{c}{Ordinals} \\
 & P & Cov & MAE & P & Cov & MAE & P & Contr & P & Contr & P & Contr \\
\midrule
containsWork & 42.0 & \textbf{29.0} & 3.7 & \textbf{49.2} & \textbf{29.0} & \textbf{2.6} & 55.0 & 33.9 & 62.5 & 40.7 & 20.0 & 25.4 \\
hasMember & 11.8 & 6.0 & 3.8 & \textbf{64.3} & \textbf{18.0} & \textbf{1.2} & 62.5 & 28.6 & 65.0 & 71.4 & 0 & 0 \\
containsAdmin & 51.8 & 14.5 & 7.3 & \textbf{78.6} & \textbf{22.0} & \textbf{1.7} & 85.7 & 87.5 & 33.3 & 10.7 & 0 & 1.8 \\
hasChild & 37.0 & \textbf{22.0} & \textbf{2.2} & \textbf{50.0} & 19.5 & 2.3 & 67.3 & 70.5 & 6.3 & 20.5 & 14.3 & 9.0 \\
hasSpouse & 26.8 & 11.0 & 1.3 & \textbf{58.1} & \textbf{12.5} & \textbf{0.5} & 75.0 & 18.6 & 43.8 & 37.2 & 63.2 & 44.2 \\
\midrule
hasZeroChild & \multicolumn{3}{c}{} & 92.3 & 18.8 & - \\
hasZeroSpouse &  \multicolumn{3}{c}{} & 71.9 & 13.7 & - \\
\bottomrule
\end{tabular}
\end{center}
\caption{Performance of CINEX-CRF in consolidating counting quantifier mentions wrt. precision (\textit{P}), coverage (\textit{Cov}) and MAE. \textit{Numt.} stands for number-related terms, \textit{Art.} for articles. Results per type show contribution (\textit{Contr}) to overall output and precision of individual types.
}
\label{tbl:consolidation-eval}
\end{table}

Table~\ref{tbl:consolidation-eval} shows the performance of CINEX-CRF, our best performing system for recognizing counting information, on the consolidation and end-to-end task. We report the results broken down per mention type, as well as in overall. 

In predicting counting quantifiers through recognizing cardinals in text, CINEX-CRF achieves 55-85\% precision. This is a considerable improvement (up to 48.9 percentage points) compared to the baseline~\cite{MirzaRDW17}.
Although the baseline yields a comparable coverage, its low precision 
suggests that it has difficulties to pick up correct context and produces some matches only by chance.

Number-related terms and articles are beneficial in improving coverage particularly for \textit{containsWork} and \textit{hasMember}, yet produce low precision results for \emph{hasChild}, possibly due to spurious indefinite articles frequently identified as counting quantifiers. 
%
%
Overall, taking compositionality as well as mention types other than cardinals into account improve both accuracy and coverage of the system, with MAE of not more than 2.6 across relations.
The performance of CINEX-CRF on predicting non-existence of objects is reported in the last two rows of Table~\ref{tbl:consolidation-eval}. We obtain a high accuracy of 92.3\% for \textit{hasChild} and 71.9\% for \textit{hasSpouse}.

\vspace{-10pt}
\subsubsection{Qualitative Analysis. } 

\begin{table}[t!]
\begin{center}
\begin{adjustbox}{width=\textwidth}
\setlength\tabcolsep{3pt}
\begin{tabular}{lllr|lr}
\toprule
 & Relation & Subject & \#$O$ & Predicted counting quantifiers \\
\midrule
\multirow{5}{*}{\rotatebox{90}{Correct}} & containsWork & The Heroes of Olympus & 5 & The Heroes of Olympus is a \under{pentalogy} of adventure... & 5 \\
 & hasMember & Siria & 2 & The music \under{duo} Siria is composed of... & 2 \\
 & containsAdmin & Gusevsky District  & 5 & ...was subdivided into \under{one} urban settlement and \under{four} rural settlements. & 5 \\
 & hasChild & Hanna Neumann & 5 & Four of her \under{five} children became mathematicians... & 5 \\
 & hasSpouse & Hannelore Schroth & 3 & Her \under{third} marriage to a lawyer produced a son... & 3 \\
\midrule
\multirow{5}{*}{\rotatebox{90}{Incorrect}} & containsWork & Scandal (TV series) & 7
 & ...this season was split into two runs, the first consisting of \under{ten} episodes. & 10 \\
 & hasMember & Ladysmith Black Mambazo  & 9 & ...Mazibuko (the eldest of the \under{six} brothers) joined Mambazo... & 6 \\
 & containsAdmin & Cottbus & 4 & Cottbus has \under{a} football team called FC Energie Cottbus... & 1 \\
 & hasChild & Barack Obama & 2 & The couple's \under{first} daughter, Malia Ann, was born on July 4, 1998. & 1 \\
 & hasSpouse & Ruth Williams Khama & 1 & ...and \under{twins} Anthony and Tshekedi were born in Bechuanaland... & 2 \\
\bottomrule
\end{tabular}
\end{adjustbox}
\end{center}
\caption{Examples of correct and incorrect predictions by CINEX-CRF.}
\label{tbl:analysis}
\end{table}

Table~\ref{tbl:analysis} lists notable examples of correct and incorrect predictions. Errors for \textit{hasMember} and \textit{hasSpouse} are sometimes caused by wrongly labelled mentions that are related instead with other relations, e.g., musical ensemble members and siblings. For some relations, understanding the fine-grained types of subject entities may help in choosing the correct context of counting quantifiers. For instance, a \textit{TV series} consists of \textit{seasons} while a specific season of the series contains \textit{episodes}. 

Notable is also the low precision of ordinals shown in Table~\ref{tbl:consolidation-eval}. A main reason is that ordinals only reliably express lower bounds (see e.g. fourth incorrect example). If one considers ordinals as correct whenever they are not higher than the true count, the reported precision scores increase from 14.3-63.2\% to 85.7-89.5\%. 


\begin{table}[t]
\centering
\begin{adjustbox}{width=\textwidth}
\begin{tabular}{ll|rr|rr|r}
\toprule
 Wikidata Subject Class &  Wikidata Property  &  P  &  Cov  &  \#Existing facts  &  \#Missing facts  &  KB increase  \\
\midrule
duo  &  has part  &  88.9 &  26.7 &  561  &  51  &  9.1\%  \\
rock band  &  has part  &  78.6 &  18.3 &  1,148  &  187  &  16.3\%  \\
band  &  has part  &  70.2 &  16.5 &  9,342  &  3,905  &  41.8\%  \\
\midrule
township of China  &  contains admin  &  100.0 &  63.0 &  7,254  &  19  &  0.3\%  \\
municipality with town privileges  &  contains admin  &  100.0 &  13.7 &  3,343  &  25  &  0.7\%  \\
amphoe (subdivision of Thailand)  &  contains admin  &  98.0 &  63.2 &  6,226  &  1,032  &  16.6\%  \\
town in China  &  contains admin  &  97.8 &  29.0 &  38,894  &  377  &  1.0\%  \\
canton of France (until 2015)  &  contains admin  &  97.2 &  38.5 &  9,191  &  189  &  2.1\%  \\
county of China  &  contains admin  &  89.5 &  35.7 &  22,401  &  236  &  1.1\%  \\
District of China  &  contains admin  &  88.9 &  35.6 &  11,828  &  170  &  1.4\%  \\
municipality of the Czech Republic  & contains admin  &  76.9 &  5.0 &  8,279  &  184  &  2.2\%  \\
\midrule
fictional human  &  child  &   100.0 &  9.1 &  327  &  141  &  43.1\%  \\
race horse  &  child  &  87.0 &  27.4 &  1,800  &  1,742  &  96.8\%  \\
mythological Greek character  &  child  &  85.7 &  21.4 &  624  &  44  &  7.1\%  \\
human biblical figure  &  child  &  66.7 &  16.7 &  274  &  42  &  15.3\%  \\
human  &  child  &  58.8 &  28.5 &  73,527  &  117,942  &  160.4\%  \\
\midrule
human  &  spouse  &  61.4 &  17.5 &  50,373  &  48,778  &  96.8\%  \\
\midrule
 \multicolumn{4}{r|}{\textbf{Total} (over all 40)}  &  \textbf{224,216}  &  \textbf{173,256}  &  \textbf{77.3\%} \\
\bottomrule
\end{tabular}
\end{adjustbox}
\caption{KB enrichment potential for 40 relations, showing only relations with accuracy (Acc) $>50\%$ and coverage (Cov) $>5\%$.}
\label{tbl:kb-enrichment}
\end{table}

\subsection{KB Enrichment Potential}
\label{subsec:kb-enrichment-40}

In this section we return to our original goal of enlarging the number of facts known to exist. 
We investigate the potential of CINEX 
on 40 relations, by focusing on the 4 previously used Wikidata properties, but looking at the up to 10 most frequent subject classes of entities using each property. For each relation, we then perform automated evaluation of CINEX as described in Section~\ref{subsec:eval-baseline}. 
In Table~\ref{tbl:kb-enrichment}, we report relations for which CINEX-CRF gave precision $>0.5$ 
and coverage $>0.05$.
%
For each relation we report the number of existing facts in Wikidata, and the existence of how many more facts we can infer from the counting quantifiers. For instance, we can derive the existence of 160.4\% more children relationships than currently stored. 
In sum, CINEX is able to identify the existence of 173K more facts than Wikidata currently knows, thus increasing the existential knowledge of Wikidata for these 40 relations by 77.3\%.

We also applied CINEX to all human entities to find out how many subjects are found to have no objects wrt.\ the \textit{hasChild} and \textit{hasSpouse} relations, finding 1,648 instances for children and 557 for spouses.
These assertions increase the existing known zero cases in Wikidata for both relations by a factor of 25.8 and 3.3, respectively.

\begin{table}[t]
\centering
\begin{adjustbox}{width=\textwidth}
\begin{tabular}{llllll}
\toprule
human  & creative works  &  admin. territorial  &  musical ensemble  &  organization  &  transport. facility  \\
\midrule
occupation  &  nominated for  &  contains settlement  &  has part  &  subsidiary  &  connecting line \\
employer  &  genre  &  contains admin. territorial  &  nominated for  &  founded by  &  adjacent station  \\
influenced by  &  cast member  &  capital of  &  record label  & -  & -  \\
award received  &  screenwriter  &  member of  &  award received  & -  & -  \\
child  &  voice actor  &  sister city  &  genre  & -  & -  \\
\bottomrule
\end{tabular}
\end{adjustbox}
\caption{Classes along with relations for which count information could be retrieved best.}
\label{tbl:across-relations}
\end{table}

\subsection{Count Information across KB Relations}
\label{sec:broad}

So far we only evaluated CINEX on four manually chosen Wikidata properties. In this section we investigate to which extent counting quantifiers are present for arbitrary relations, and to which extent they can be extracted by CINEX.

To this end, we collected all Wikidata properties that were interesting, i.e., were not asserted to be single-value\footnote{Properties having the  constraint \url{https://www.wikidata.org/wiki/Q19474404}.}, had a functionality degree
$(\#\textit{subjects}/\#\textit{triples})$ of less than 0.98~\cite{galarraga2015fast}, and were used by at least 500 subjects, obtaining 267 properties in total.
For each of these properties, we identified the 10 most frequent entity classes used as subjects, resulting in a total of 2,474 relations. 
For each relation, we then performed automated evaluation of CINEX as described in Section~\ref{subsec:eval-baseline}, finding 110 relations for which CINEX gave precision $> 50\%$ and coverage $>5\%$. 


Among the frequent classes (grouped by theme) of subjects for which we can mine counting quantifiers from the corresponding Wikipedia pages are: \textit{human} (including twin, fictional human, biblical figure and mythological Greek character), \textit{creative works} (e.g., film, television series), \textit{administrative territorial entity} (e.g., country, municipality), \textit{musical ensemble} (e.g., band, duo), \textit{organization} (e.g., business enterprise, nonprofit organization) and \textit{transportation facility} (e.g., metro station, train station). We show in Table~\ref{tbl:across-relations} the top 5 Wikidata properties 
for each mentioned subject type. Other notable relations include: \textit{$<$battle, participant$>$, $<$human spaceflight, crew member$>$ and $<$star, child astronomical body$>$}.

In terms of KB enrichment, CINEX was able to extract a total of 
851K
counting quantifier facts, which in turn state the existence of 2.5M facts not yet asserted for these 110 $<$\textit{Wikidata class}, \textit{Wikidata property}$>$ pairs.
These existential facts, provided on Github, increase the number of facts known to exist for these relations by 28.3\%.

\section{Related Work}
\label{sec:relatedwork}

Knowledge bases have seen a rise of attention in recent years. Aside from a few manual efforts like Wikidata, the construction of these knowledge bases is usually done via automated information extraction, focusing either on structured data (DBpedia~\cite{dbpedia}, YAGO~\cite{suchanek2007yago}), or on unstructured contents from the web. For the latter, directions include extracting arbitrary facts without predefined schema, called Open IE~\cite{Mausam2012,delcorro2013clausie,NELL:Mitchell2015}, and extracting triples based on well-defined knowledge base relations~\cite{surdeanu-EtAl:2012:EMNLP-CoNLL,koch-EtAl:2014:EMNLP2014,DBLP:conf/icwsm/PalomaresAKR16}, in which the distant supervision approach is widely used~\cite{distant-first,Mintz:ACL2009,Suchanek:WWW2009}. 
The idea of distant supervision is to use facts from an existing KB in order to label sentences as positive/negative training samples, depending on whether the entities from the existing facts occur in them or not. 
A major challenge for distant supervision is knowledge base incompleteness: If the KB used for labeling the training data misses facts, candidates may wrongly be classified as negative samples, reducing the quality of the learning process. Approaches to mitigate this effect include heavily under-sampling the negative evidence~\cite{undersample1,surdeanu-EtAl:2012:EMNLP-CoNLL}, to learn only from positive samples~\cite{bonanmin}, or to use heuristics in selecting negative samples~\cite{galarraga2015fast,knowledgevault}, yet these do not help with potentially wrong seed counts. 

Most works on information extraction focus on relations that link entities, like $\langle \textit{Trump,}$ $\textit{presidentOf, USA} \rangle$, or that store String or measurement values. Counting quantifiers have received comparably little attention. 
Numbers, a major construct for expressing counts, were investigated mostly in the context of temporal information, e.g.\ to enrich facts with timestamps/durations~\cite{ling2010temporal,strotgen2010heideltime}, or in the context of quantities and measures like \emph{$\langle$MtEverest, height, 8848mt$\rangle$}~\cite{madaan2016numerical,yusra-CIKM-2016,ISWC-2016-numeric-values,mausam2017bootstrapping}. 
In contrast, terms that express counting quantifiers are either extracted incorrectly by state-of-the-art Open-IE systems, or not at all. While NELL, for instance, knows 13 relations about the number of casualties and injuries in disasters, they all contain only seed facts and no learned facts. In~\cite{MirzaRDW17}, which we use as baseline for our experiments, we have proposed a single-stage process for identifying numbers that express relation counts. Yet, we there only consider explicit cardinals and do not tackle training data incompleteness nor compositionality, thus achieving 
only moderate precision and coverage.


While a few counting qualifier predicates such as \emph{number of children},
\emph{number of seasons} (of a TV series)
or \emph{number of households} (of a territory) already exist in Wikidata, it should be noted that a proper interpretation of counting quantifiers requires to go beyond the standard open-world assumption of the Semantic Web, as they allow to infer negative information. Appropriate models require to combine open-world and closed-world reasoning, as does for instance the local closed-world assumption~\cite{Denecker:Calabuig-logical_theory_partial_databases:tods:10,darari:ISWC:2013}.

\section{Conclusions}

\renewcommand{\baselinestretch}{1.1}
\parbox{\textwidth}{
We have proposed to enrich KBs with counting quantifiers, and discussed the challenges that set counting quantifier extraction apart from standard information extraction. In particular, we showed that it is imperative to consider the compositionality of counts, and their expression in non-numeric form. We have shown that our system, CINEX, can extract counting quantifiers with 60\% average precision on five relations, and when applied to a large set of relations, it is possible to extend the number of facts known to exist in 110 of them by 28\%. We believe that the extraction of counting quantifiers opens interesting avenues for tasks such as question answering, information extraction or KB curation. Our data and code are available at
\mbox{\ \ \ \ \ \ \ \ }\linebreak 
\footnotesize{\url{https://github.com/paramitamirza/CINEX}}.
}

\renewcommand{\baselinestretch}{1.0}

\bibliographystyle{plain}
{
\bibliography{refs}
}






\end{document}